%% file: 00_nsnmf.tex
\documentclass[letterpaper]{article}
\pdfoutput=1
\usepackage[margin=5pt]{subcaption}
\usepackage{comment}
\usepackage{aaai}
\usepackage{times}
\usepackage{helvet}
\usepackage{courier}
\usepackage[fleqn]{amsmath}
\usepackage{amssymb,bm}
\usepackage{algorithm}
\usepackage{algorithmic}
\usepackage{color}
\usepackage{xspace}
\usepackage{multirow}
\usepackage[pdftex]{graphicx}

\newcommand{\nagc}{NAGC\xspace}
\newcommand{\proposal}{our method\xspace}
\newcommand{\nlf}{a non-linear projection function\xspace}
\newcommand{\cls}{cluster assignment\xspace}

\newcommand{\bmX}{\bm{X}}
\newcommand{\bmS}{\bm{S}}
\newcommand{\bmU}{\bm{U}}
\newcommand{\bmV}{\bm{V}}
\newcommand{\bmA}{\bm{A}}
\newcommand{\bmZ}{\bm{Z}}
\newcommand{\bmW}{\bm{W}}
\newcommand{\trans}{^{\top}}
\newcommand{\inmathbb}[2]{\in \mathbb{R}_+^{#1 \times #2}}
\newcommand{\inmathbbr}[2]{\in \mathbb{R}^{#1 \times #2}}
\newcommand{\bmH}{\bm{H}}
\newcommand{\bma}{\bm{\alpha}}
\newcommand{\bmb}{\bm{\beta}}
\newcommand{\bmg}{\bm{\gamma}}
\newcommand{\F}{\mathcal{F}}
\newcommand{\Tr}{{\rm Tr}}
\DeclareMathOperator{\argmax}{argmax}

\begin{document}
%
\makeatletter



\title{
Non-linear Attributed Graph Clustering by\\ Symmetric NMF with PU Learning}

\author{Seiji Maekawa$^1$, Koh Takeuchi$^2$, Makoto Onizuka$^1$\\
$^1$Graduate School of Information Science and Technology, Osaka University, 1--5 Yamadaoka, Suita, Osaka, Japan\\
$^2$NTT Communication Science Laboratories, 2--4 Hikaridai, Seika, Soraku, Kyoto, Japan\\
$^1$\{maekawa.seiji,onizuka\}@ist.osaka-u.ac.jp, $^2$koh.t@acm.org
}

\maketitle
\begin{abstract}
\begin{quote}
We consider the clustering problem of attributed graphs. 
Our challenge is how we can design an effective and efficient clustering method that precisely captures the hidden relationship between the topology and the attributes in real-world graphs.
We propose Non-linear Attributed Graph Clustering by Symmetric Non-negative Matrix Factorization with Positive Unlabeled Learning.
The features of our method are three holds. 
1) it learns \nlf between the different \cls{s} of the topology and the attributes of graphs so as to capture the complicated relationship between the topology and the attributes in real-world graphs,
2) it leverages the positive unlabeled learning~\cite{liu2003building} to take the effect of partially observed positive edges into the cluster assignment, and
3) it achieves efficient computational complexity, $O((n^2+mn)kt)$, where $n$ is the vertex size, $m$ is the attribute size, $k$ is the number of clusters, and $t$ is the number of iterations for learning the \cls.
We conducted experiments extensively for various clustering methods with various real datasets to validate that our method outperforms the former clustering methods regarding the clustering quality. 
\end{quote}
\end{abstract}

\input{01_intro}
\input{02_nmf}

\input{03_prop}
\input{04_experiment}
\input{06_related}
\input{05_conclusion}

\bibliographystyle{aaai}
{\small
\bibliography{10_aaai}
}

\end{document}

%% file: 01_intro.tex
\section{Introduction}
\label{sec:introduction}
Graph is a fundamental data structure for representing vertices and their relationships. 
Graph data appear everywhere in many application domains, such as web graph~\cite{flake2002self}, social network~\cite{fortunato2010community}, 
protein complexes~\cite{brohee2006evaluation}, 
traffic planning~\cite{george2007spatio},
computer vision~\cite{jain2016structural},
and gene expressions~\cite{ben1999clustering,kulis2009semi}.
The authors of ~\cite{Sahu2017ubiquity} conducted an online survey and 
showed that graph database is becoming increasingly prevalent across many application domains and, in particular, 
the graph clustering is the most widely used technique in machine learning and data mining fields.

Graphs in the real world usually have attributes on vertices.
Actually, the graph databases support attributed graphs~\cite{Francis2018Cypher,Sevenich2016}.
However, most of the graph clustering techniques~\cite{newman2006modularity,xu2007scan,karypis1998multilevelk} do not leverage the attributes of vertices since their design is limited to simple graphs without having attributes.
Therefore, these techniques can not extract precise clusters without leveraging the attributes.

There are emerging researches that tackle the clustering problem for attributed graphs~\cite{huang2017joint,xu2012model,Akoglu2012pics,parimala2015sans,zhou2009graph,zhou2010clustering}. 
Despite the considerable improvements made by the existing methods, they have not fully leveraged the virtue of attributed real-world graphs.
That is, there are two missing aspects of the attributed graphs we should consider for designing effective clustering methods.
First, the topology and the attributes of real-world graphs have a complicated relationship with each other, that is, they would have different cluster structures in general,
because the topology and attributes are obtained from different viewpoints of the similarity of vertices.
Second, typical graphs usually have a subset of positive edges,
since real-world graphs follow the open world assumption, that is ``absence of information is interpreted as unknown information, not as negative"~\cite{Keet2013}.
For example, a social graph may not reflect precisely the social connections in the real world:
we can only observe positive connections between people such as ``likes" and ``friendships", but cannot observe negative ones~\cite{hsieh2015pu}. 
In addition, there is a possibility of missing positive edges among pairs of vertices where no edges were observed in the graph.




We take the above two aspects into account and propose \nagc, Non-linear Attribute Graph Clustering by Symmetric Non-negative Matrix Factorization with Positive Unlabeled Learning.
To achieve high clustering quality, 
1) our method flexibly captures the complex relationship among the topology and the attributes by learning \nlf among their different \cls{s}, and
2) \proposal leverages PU learning~\cite{elkan2008learning,liu2003building,hsieh2015pu} to take the effect of partial positive edges and no edge observations into the cluster assignment.
To the best of our knowledge, our method is the first method that applies the idea of PU learning to the graph clustering.
Our method can precisely capture clustering results by revealing the relationship between the topology and the attributes in real-world graphs.
As for the efficiency, we carefully design the learning model of our method so that its cost does not contain the quadratic effect of the number of attributes. 
Thus, our method is scalable to the number of attributes, which is usually large in real-world graphs (See the statistics of the graphs in Table~\ref{tb: dataset} in the experiment section).


We extensively made experiments for various clustering methods over various real datasets with ground truth. We compared our method with the former clustering methods in terms of the clustering quality and efficiency. 
We also evaluate the effectiveness of PU learning and the effect of the hyperparameters.
With these experiments, we confirm that our method outperforms the existing methods in terms of the clustering quality by leveraging the different but related cluster structures among the topology and the attributes.
We also confirm that our method is stable against the hyperparameter selection. 

The rest of this paper is organized as follows.
We introduce fundamental techniques for our method, Non-negative Matrix Factorization, Symmetric Non-negative Matrix Factorization, and Biased Matrix Completion in the preliminary section.
We propose our method in the section of non-linear attributed graph clustering.
The experiment section gives the purpose and results of the evaluations.
The related work section addresses the details of the related work and we conclude this paper at the last section.

%% file: 02_nmf.tex
\section{Preliminaries}
\label{sec:preliminaries}

{\bf Notation:} We denote a matrix and its $i$-th row vector as upper boldface $\bmX$ and under boldface $\bm{x}_i$.
The set of non-negative real numbers is $\mathbb{R}_+$.
We denote a graph $G=(V,E)$ comprising a set of vertices $V=\{1,2,\dots,n\}$ and edges $E=\{( i,j)\}\subseteq [n] \times [n]$.
We construct an weighted adjacency matrix $\bmS \inmathbb{n}{n}$ from $G$, where $s_{i,j}$ is set to a positive value if there is a edge between two vertices $i$ and $j$ or set to $0$ otherwise.
We denote a non-negative attribute matrix $\bmX \inmathbb{n}{m}$ that represents $n$ vertices with $m$ attributes. 
$\| \cdot \|_\F$ and $\|\cdot\|_*$ are Frobenius norm and the nuclear norm.
We use $\odot$ and $\oslash$ to denote the element-wise multiplication and the element-wise division.


\subsection{Non-negative Matrix Factorization}
\label{sec:nmf}
Given a number of clusters $k_1 \ll \min\{m,n\}$, we suppose a non-negative factor matrix for a cluster assignment $\bmU \inmathbb{n}{k_1}$
and an attribute factor matrix $\bmV \inmathbb{m}{k_1}$. 
Non-negative Matrix Factorization~(NMF)~\cite{lee1999learning} estimates local optimal parameters $\bmU$ and $\bmV$  by minimizing a non-convex loss between $\bmX$ and its approximation $\bmU\bmV\trans$.
\begin{align}
\label{eq: nmf}
\min_{\bmU,\bmV \ge0} \| \bmX-\bmU\bmV\trans\|^2_\F
\end{align}
Thanks to the non-negative constraint, we can obtain a clustering result of $i$-th vertex by choosing the index that has the largest value in the $i$-th row vector of $\bmU$.

Non-negative Matrix Tri-Factorization~\cite{ding2006orthogonal} is a novel extension of NMF that supposes different numbers of clusters for vertices and attributes. This method introduces a transfer matrix $\bmH \inmathbb{k_1}{k_2}$ that can represent the relationship between the vertex and attribute clusters, where $k_2$ denotes the number of clusters for the attributes.
They decompose the attribute matrix with learning $\bmH$.
\begin{align}
\label{eq: tri-factorization}
\min_{\bmU,\bmV,\bmH \ge0} \| \bmX-\bmU\bmH\bmV\trans\|^2_\F
\end{align}
where $\bmV \inmathbb{m}{k_2}$ is an attribute factor matrix with $k_2$ factors.
We can put an orthogonal constraint on factor matrices to get precise clusters.
This method is limited to consider linear relationships among clusters of vertices and attributes.

\subsection{Symmetric Non-negative Matrix Factorization}
The goal of graph clustering is to find a partition of vertices in a graph where the similarity between vertices is high within the same cluster and low across different clusters.
To capture such a cluster structure embedded in a graph, Kuang et. al proposed Symmetric Non-negative Matrix Factorization (SNMF)~\cite{kuang2015symnmf,kuang2012symmetric}, and showed an interesting relationship among SNMF and graph clustering methods including the spectral clustering~\cite{ng2002spectral}.

SNMF estimates a cluster assignment matrix $\bmU$ by minimizing a non-convex loss function that uses $\bmS$ as input:
\begin{align}
\label{eq: symnmf}
\min_{\bmU \ge0}\|\bmS-\bmU\bmU\trans\|^2_\F
\end{align}
In the same manner as NMF, we can obtain a clustering result by assigning $i$-th vertex to the $k_1'$-th cluster that has the largest value in $\bm{u}_i$, that means $k_1' = \argmax_{l} \{u_{i,l}\mid l=(1,\dots,k)\}$.

\subsection{Biased Matrix Completion}
\label{sec:biasmc}
Hsieh et. al~\cite{hsieh2015pu} considered a matrix completion problem when only a subset of positive relationships is observed, such as recommender systems and social networks where only ``likes" or ``friendships" are observed.
The problem is an instance of PU (positive-unlabeled) learning~\cite{elkan2008learning,liu2003building}, i.e. learning from only positive and unlabeled examples that has been studied in the classification problems.
They introduced the $\rho$-weighted loss for a bipartite graph $G'=(V',E')$ comprising a set of vertices $V'=\{\{1,2,\dots,n\}, \{1,2,\dots,m\}\}$ and edges $E='\{(i,j)\}\subseteq [n] \times [m]$: 
\begin{align}
\label{eq: weighted loss}
\ell_{\rho}(z_{i,j}) = \rho{1}_{(i,j)\in E'}(z_{i,j}-1)^2 + (1-\rho){1}_{(i,j) \not\in E'}z_{i,j}^2
\end{align}
where $\rho=[0,1]$, ${1}_{(i,j)\in E'}(\cdot)$, and ${1}_{(i,j)\not\in E'}(\cdot)$ are a bias weight, an indicator function for positive edges, and an indicator function for unlabeled edges, respectively.
This loss can change a weight for reconstruction errors among positive and unlabeled edges.
When we set $\rho = 0.5$, it treats the positive and unlabeled entities equally. 
With this loss, they proposed a biased matrix completion as:
\begin{align}
\label{eq: objective of PU}
\min_{\bmZ: \|{\bm Z}\|_* \leq \lambda}  \sum_{(i,j)\in E'} \rho(z_{i,j}-1)^2+\sum_{(i,j) \not\in E'}(1-\rho)z^2_{i,j}
\end{align}
where $\lambda \geq 0$ is a hyperparameter.
To the best of our knowledge, no method based on SNMF has employed the biased formulation of matrix completion problems.


%% file: 03_prop.tex
\input{table/variables.tex}

\begin{figure}[t]
  \centering
  \includegraphics[trim={0mm 0mm 0mm 0},width=8cm]{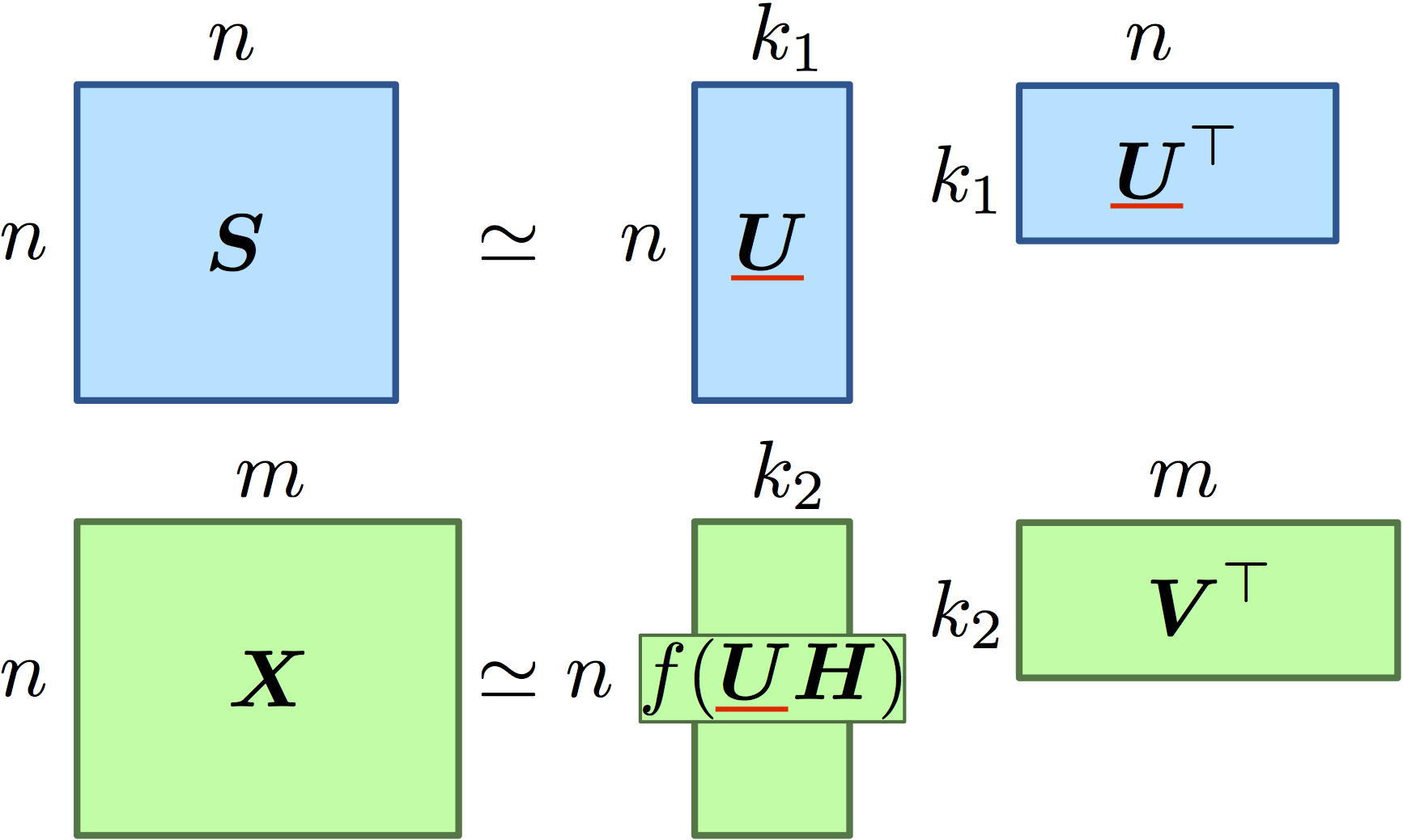}
  \caption{
  Illustration of \nagc. $\bmS$ and $\bmX$ are an adjacency matrix and an attributed matrix, respectively. $\bmU$, $\bmV$, and $\bmH$ denote a cluster assignment, an attribute factor, and a cluster assignment transfer matrices, respectively. $f$ is a non-linear activation function. \nagc merges different cluster structures among $\bmS$ and $\bmX$ by supposing a shared parameter $\bmU$.
  }
  \label{model_figure}
\end{figure}

\section{Non-linear Attribute Graph Clustering}
\label{sec:proposal}
In this section, we propose a variant of NMF that can find a reasonable cluster assignment by considering the complex relationship between the topology and the attributes. 
We also attempt to mitigate the serious problem of the partial positive edges that prevents us from capturing a precise assignment by leveraging the idea of the biased matrix completion.
Table \ref{tb: variables} lists the main symbols and their definitions.

Our method jointly decomposes the adjacency matrix $\bmS$ and the attribute matrix $\bmX$ into factor matrices with learning \nlf.
This function can transfer a cluster assignment extracted from the adjacency matrix to that from the attribute matrix.
Here, we define our method as a minimization problem of a non-convex loss.
\begin{align}
\label{eq: objective of proposal}
\min_{\bmU, \bmV, \bmH \ge 0} \mathcal{L}_{\rho}(\bmS-\bmU\bmU\trans) + \frac{\lambda}{2}\|\bmX-f(\bmU\bmH)\bmV\trans\|^2_\F 
\end{align}
where $f$ denotes an element-wise non-linear activate function.
We use $\mathcal{L}_{\rho}(\bmZ)$ to denote an approximation error of the adjacency matrix $\bmS$ with the $\rho$-weighted loss.
\begin{align}
    \mathcal{L}_{\rho}(\bmZ) = \sum_{(i,j)\in E}\rho(z_{i,j}-1)^2+(1-\rho)\sum_{(i,j) \not\in E}z^2_{i,j}
\end{align}
We employ the sigmoid function as a non-linear activation function in this paper: $f(x) = \frac{1}{1+e^{-x}}$. 
This choice can be generalized to any non-linear functions.
Note that, in the second term of Eq.~\eqref{eq: objective of proposal}, $f(\bmU\bmH)$ plays the role of the cluster assignment matrix for the attribute matrix. By transforming $\bmU$ with $f$ and $\bmH$, our method enables to capture the complex relationship among the topology and the attributes.
Our method works as SNMF when $\lambda=0$ and $\rho=0.5$.
We illustrate our proposed method in Figure \ref{model_figure}.

Since our loss is non-convex for $\bmU$, $\bmV$, and $\bmH$, we derive a parameter estimation procedure that alternatively updates each parameter by utilizing the method of Lagrange multipliers~\cite{ding2006orthogonal}. 
Following the standard theory of constrained optimization, we introduce Lagragian multipliers $\bma \inmathbbr{n}{k_1}$, $\bmb \inmathbbr{m}{k_2}$, and $\bmg \inmathbbr{k_1}{k_2}$ for the non-negative constraints $\bmU,\bmV,\bmH \geq 0$. We define the Lagrangian function of our proposed method as:
\begin{align}
\label{eq: loss function}
&\mathcal{L}(\bmU,\bmV,\bmH;\bma,\bmb,\bmg)\nonumber\\
&= \mathcal{L}_{\rho}(\bmS-\bmU\bmU\trans) + \frac{\lambda}{2}\|\bmX-f(\bmU\bmH)\bmV\trans\|^2_\F \nonumber\\
&\quad + \Tr(\bma\trans\bmU)+\Tr(\bmb\trans\bmV)+\Tr(\bmg\trans\bmH)
\end{align}

For each parameter $\bmU,\bmV$, and $\bmH$, we derive partial differences of the Lagrangian function.
\begin{align}
\label{eq: gradient of U}
\frac{\partial \mathcal{L}}{\partial \bmU}& = -2\rho\bmS\bmU -\lambda \{(\bmX \bmV) \odot f'(\bmU \bmH)\} \bmH\trans\nonumber\\ 
&\quad + 2\rho(\bmU\bmU\trans\odot\bmW)\bm{U}+ 2(1-\rho)(\bmU\bmU\trans\odot\bmW')\bm{U} \nonumber\\
&\quad +\lambda [\{f(\bmU \bmH) \bmV\trans\bmV\} \odot f'(\bmU \bmH)] \bmH\trans + \bma\\
\label{eq: gradient of V}
\frac{\partial \mathcal{L}}{\partial \bmV} 
&= -\lambda \bmX\trans f(\bmU \bmH) + \lambda \bmV f(\bmU\bmH)\trans f(\bmU \bmH)+\bmb\\
\label{eq: gradient of H}
\frac{\partial \mathcal{L}}{\partial \bmH} &= 
- \lambda \bmU\trans \{f'(\bmU\bmH)\odot(\bmX\bmV)\} \nonumber\\
&\quad + \lambda \bmU\trans \{f'(\bmU\bmH)\odot f(\bmU\bmH)\}\bmV\trans \bmV + \bmg
\end{align}
where $\bmW \inmathbb{n}{n}$ is a mask matrix whose elements are set as $w_{i,j}=1$ if $s_{i,j}\ne0$ or $w_{i,j}=0$ otherwise and $\bmW'= 1 - \bmW$.
The KKT complementarity conditions for the non-negative constraints of parameters are:
\begin{align}
    &\bma\odot \bmU={0}, \bmb\odot \bmV={0}, \bmg\odot \bmH={0} \\
    &\frac{\partial \mathcal{L}}{\partial \bmU}=0, \frac{\partial \mathcal{L}}{\partial \bmV}=0, \frac{\partial \mathcal{L}}{\partial \bmH}=0
\end{align}
By satisfying these conditions, we can derive multiplicative update rules for each parameter.
\begin{align}
\label{eq: update of U}
\bmU& \leftarrow \bmU\odot [2\rho\bmS\bmU
+\lambda \{(\bmX \bmV)\odot f'(\bmU \bmH)\} \bmH\trans] \oslash\nonumber\\
&\quad [2\rho(\bmU\bmU\trans\odot\bmW)\bmU + 2(1-\rho)(\bmU\bmU\trans\odot\bmW')\bmU \nonumber\\
&\quad+\lambda \{(f(\bmU\bmH)\bmV\trans\bmV) 
\odot f'(\bmU\bmH)\} \bmH\trans]\\
\label{eq: update of V}
\bmV& \leftarrow \bmV \odot \{\bmX\trans f(\bmU\bmH)\}\oslash\{\bmV f(\bmU\bmH)\trans f(\bmU\bmH)\}\\
\label{eq: update of H}
\bmH& \leftarrow \bmH\odot [\bmU\trans \{f'(\bmU\bmH)\odot(\bmX\bmV)\}]\nonumber\\
&\quad \oslash [\bmU^T \{f'(\bmU\bmH)\odot f(\bmU\bmH)\}\bmV\trans\bmV]
\end{align}
Our loss is convex with respect to $\bmV$ and $\bmH$, however, as mentioned in~\cite{kuang2012symmetric}, the loss is a fourth-order non-convex function with respect to $\bmU$. 
That means, it is difficult to guarantee the monotonic convergence of our parameter estimation method; thus we expect a good convergence property that every limit point is a stationary point. 
We show the parameter estimation and clustering algorithm for our proposed method in Algorithm~\ref{al: njnmf}.

Since non-convex minimization problems have multiple local minima, we employ two popular methodologies for initializing the parameters $\bmU,\bmV$ and $\bmH$. One is to use random values that is a typical way for matrix factorization. The other is to put the result of k-means as initial values of $\bmU$ and $\bmV$. 
$\bmH$ is initialized by random values in both cases because there is no corresponding information to $\bmH$. 

\subsection{Computational complexity}
Here, the computational complexity is discussed. 
We stop our algorithm at $t$ iterations, then the overall cost for SNMF is $O(n^2kt)$~\cite{kuang2015symnmf,kuang2012symmetric}. 
JWNMF that is the state-of-the-art method for the attributed graph clustering needs $O((n^2 +m^2 + mn)kt)$~\cite{huang2017joint} because it computes extra parameters for the joint factorization and parameter selections. 
Our proposed method does not need to calculate $\bm A$ in Eq.(\ref{eq: objective of jwnmf}) thus the overall cost for updating rules is equal to $O((n^2+mn)kt)$ where $k = \max(k_1,k_2)$ and $k \ll n$.
Therefore, our proposed method is much faster than JWNMF and almost the same as the original SNMF when $m \ll n$.

\begin{algorithm}[t]
 \begin{algorithmic}[1]
  \REQUIRE $\bmS,\bmX, k_1, k_2, \lambda, t$
  \ENSURE clustering result ${C}$
  \STATE Preprocess: $\bmS, \bmX$ 
  \STATE Initialize: {$\bmU, \bmV, \bmH$}
  \WHILE{$t' < t$}
    \STATE \# alternatively update parameters
    \STATE ${\bmU}^{(t'+1)} \leftarrow$ update $({\bmU}^{(t')})$ ~by Eq.~\eqref{eq: update of U}
    \STATE ${\bmV}^{(t'+1)} \leftarrow$ update $({\bmV}^{(t')})$ ~by Eq.~\eqref{eq: update of V}
    \STATE ${\bmH}^{(t'+1)} \leftarrow$ update $({\bmH}^{(t')})$ ~by Eq.~\eqref{eq: update of H}
  \ENDWHILE
  \WHILE{$n' < n$}
    \STATE \# assign each vertex to the clusters
    \STATE ${c_{n'}} \leftarrow \argmax_{l}\{u_{n',l}\mid l=(1, \dots, k)\}$ 
  \ENDWHILE
 \end{algorithmic}
 \caption{\nagc algorithm}\label{al: njnmf}
\end{algorithm} 

%% file: table/variables.tex
\begin{table}[t]
  \caption{Definition of main symbols.}
  \label{tb: variables}
\setlength{\tabcolsep}{5pt}
  \begin{center}
	\begin{tabular}{|l|l|} \hline
	Variable & Explanation \\ \hline\hline
    $\bmS \inmathbb{n}{n}$ &  adjacency matrix \\ \hline
    $\bmX \inmathbb{n}{m}$ &  attribute matrix \\ \hline
    $\bmU \inmathbb{n}{k_1}$ &  cluster assignment matrix \\ \hline
    $\bmV \inmathbb{m}{k_2}$ &  attribute factor matrix \\ \hline
    $\bmH \inmathbb{k_1}{k_2}$ &  cluster assignment tansfer matrix \\ \hline
    $\bmW \inmathbb{n}{n}$ &  mask matrix of $\bmS$ \\ \hline
    $k_1 \in \mathbb{N}$ &  number of clusters \\ \hline
    $k_2 \in \mathbb{N}$ &  number of clusters for attributes \\ \hline
    \multirow{2}{*}{$\lambda \ge 0$} &  balancing parameter between \\ 
    &  the topology and the attributes \\ \hline
    $ \rho = [0,1] $ & bias weight for $\bmS$  \\ \hline
    $t \in \mathbb{N}$ &  number of iterations  \\ \hline
 	\end{tabular}
 \end{center}
 \hspace{-5cm}
\end{table}

%% file: 04_experiment.tex
\section{Experiments}
\label{sec:experiments}
The first goal of the experiments is to show \proposal\footnote{The source code of our method, \nagc, is available at https://github.com/seijimaekawa/NAGC.} outperforms JWNMF~\cite{huang2017joint}, the state-of-the-art method for attributed graph clustering. 
Let us start from the visualized clustering results\footnote{
We utilize Gephi for visualization.
Gephi is limited to place the vertices based only on the graph topology and ignores the effect of the attributes.}
for WebKB dataset (see the following subsection for the details of the dataset), which is shown in Fig.~\ref{fg: visualization}. 
We observe that both two methods successfully assign the clusters to the vertices located near the center.
However, JWNMF fails to assign the precise clusters to many surrounding vertices (``x" indicates the assignment error in the figure). 
We investigated the reason and found that JWNMF ignores the effect of most attributes and 
this prevents JWNMF to learn the precise cluster assignment.
This is caused by the limitation of JWNMF: it only learns a simple relationship between the topology and the attributes to have the same clustering structure.
In contrast, \proposal successfully assigns the precise clusters.
It learns a non-linear projection function among the different cluster assignments of the topology and the attributes.

\input{fig/0_visual.tex}

The WebKB dataset is not a special case \proposal works well.
Our second goal of the experiments is to evaluate the clustering quality and efficiency of \proposal for various datasets.
We evaluate \proposal with the former methods, JWNMF and BAGC~\cite{xu2012model}, those are designed for attributed graphs. 
We also evaluate simple graph clustering methods without using attributes, METIS~\cite{metis1998} and SNMF, and attribute-based clustering methods, NMF and k-means, so that how much only the topology or attributes of the graphs contribute to the clustering quality.
We used publicly available codes for the existing methods in our experiments.
To investigate the details of the quality improvement achieved by \proposal,
we also evaluate the effectiveness of PU learning and the effect of the hyperparameters.
We perform five restarts for each method and calculate the average and standard deviation of the results.

\subsection{Datasets}
Four real-world datasets with the ground truth are employed in our experiments.
WebKB\footnote{http://linqs.cs.umd.edu/projects//projects/lbc/index.html\label{footnote webKB}} is the web graph of four universities: the label for a vertex indicates the owner university of the page. The attributes of a vertex represent the words appeared in the page.
Citeseer and Cora (see also footnote\ref{footnote webKB} for detail) are citation networks.
The label for a vertex corresponds to a research field of the paper.
The attributes of a vertex consist of the words appeared in the paper. 
Polblog\footnote{http://www-personal.umich.edu/\~{}mejn/netdata/} is a network of hyperlinks between blogs on US politics: the label of a vertex indicates whether the blog is liberal or conservative. The attributes of a vertex represent the sources of the blogs.
Table~\ref{tb: dataset} summarizes the characteristics of the four datasets. 
The density column in the table indicates (\# of observed edges)/(\# of possible edges), that is ${{|{E}|}/{\ n^2}}$.
\input{table/dataset}

\subsection{Measurements}
We utilize the Adjusted Rand Index (ARI) \cite{yeung2001details}, which is a typical measurement used for assessing the clustering quality with ground truth labels.
Generally, higher ARI indicates better clustering results.
In addition to ARI, we also employ the modularity \cite{newman2006modularity} and average entropy for cluster validation with respect to the topological aspect and attribute aspect of clusters, respectively.
Intuitively, higher modularity indicates there are dense connections in the same cluster but sparse connections between different clusters.
Lower average entropy indicates there are similar attribute values in the same cluster but dissimilar attribute values between different clusters.

\subsection{Parameter Settings}
In our experiments, we made a grid search for each dataset to select parameters, $\lambda$, $k_2$, and $\rho$, those give the best performance.
$\lambda$ is chosen from the set $\{10^{-10}, 10^{-8}, 10^{-7}, 10^{-6}, 10^{-5}, 10^{-4}, 10^{-3},\allowbreak 10^{-2},\allowbreak 0.1,\allowbreak 1, 10, 100\}$
by following the settings used in \cite{huang2017joint}. 
The model does not work well when $\lambda>100$.
We set $k_1$ at the number of ground truth labels for each dataset.
$k_2$ is chosen from the set $\{k_1,5,7,10,15,20\}$ so that we can learn the model more precisely than the number of clusters.
$\rho$ is chosen from the set $\{0.5,0.55,0.75,0.95,0.995\}$.
To mitigate the different scales between $\bmS$ and $\bmX$, we normalize $\bmS$ by 
multiplying each element of $\bmS$ with $\frac{|\bmX|}{|\bmS|}$.
The number of the iterations $t$ is fixed at $100$ in all the experiments.

\subsection{Clustering quality}
\input{table/ari}
\input{table/modularity}
Table~\ref{tb: ari} shows the results of evaluating the clustering quality by using the average and standard deviation of ARI.
We confirmed that \proposal initialized by k-means results (Prop.) outperforms all the competing methods for all datasets.
This result validates the effectiveness of the non-linear projection and PU learning to the clustering quality.
The results of our method without PU learning (Prop. (w/o PU)) clarify the advantage of the non-linear projection: it performs better than the competing methods except for Cora dataset.
In addition, the initialization by k-means result always improves the performance, see the gain from Prop.$^{*}$ (initialized by random values) to Prop. (initialized by k-means result).

JWNMF took the second place for WebKB dataset but resulted in poor performance on other datasets.
METIS, that is a graph clustering method, achieved the second or third places for WebKB, Cora, and polblog datasets.
In contrast, for Citeseer dataset, NMF and k-means, those are attribute-based clustering methods, took the second and third places, respectively.
These results indicate that either the topology or attributes of the graphs contribute largely to the clustering quality, but it is more effective to combine both of them.



To investigate more on the difficulty of the attributed graph clustering, 
we show that the topology and the attributes of real-world graphs have different cluster structures.
Table~\ref{tb: modularity} gives the modularity and the average entropy for the clustering result of WebKB dataset.
We also include the ARI in the table for comparison, which is the same result as in Table~\ref{tb: ari}.
Our method achieves the highest ARI but does not achieve either the best modularity or the best average entropy.
This result implies that, when we design an effective clustering method for attributed graphs, we should not optimize the model only to either the topology or the attributes. 
Instead, we need to take both effects of the topology and attributes.

As for BAGC, the ARI is the lowest among all the methods in all the datasets. 
It achieves better entropy than other attributed graph clustering methods do (our method and JWNMF),
however, it fails to learn the model fully from the graph topology.
We also note on the evaluation of PAICAN~\cite{bojchevski2018bayesian},
one of the latest methods designed for outlier detection and clustering of attributed graphs. 
Despite an extensive search we made on the hyperparameters, 
PAICAN resulted in poor performance in our experiments, so we exclude its results from Table~\ref{tb: ari}.
\footnote{The clustering quality reported in the PAICAN paper is high because the clusters are obtained only from regular vertices after removing the outliers.
This setting is different from other papers.}


\subsection{Hyperparameter Discussion}
\input{fig/0_lambda.tex}
\input{fig/0_k2.tex}
\input{table/q-ari}

We discuss the effect of the hyperparameters of \proposal.
We show the results of the variation of our method initialized by k-means result.
Fig.~\ref{fg: lambda} shows the effect of $\lambda$ to the clustering results. 
Other parameters are fixed at the values when ARI becomes highest for each $\lambda$.
There is a peak in each dataset ($\lambda=10^{-4}$ on WebKB and $\lambda=10^{-2}$ on Cora) which indicates that the effect to the model is well balanced by $\lambda$ between the topology and the attributes.


The effect of $k_2$ to ARI is shown in Fig.~\ref{fg: k2}.
Fig.~\ref{fg: WebKB_k2} shows that ARI slightly increases when $k_2$ increases. 
ARI of the WebKB is enough high (almost $1.0$) when $k_2=20$.
Fig.~\ref{fg: cora_k2} shows that there is a peak of ARI on Cora when $\rho=0.95$ and $k_2=10$.
ARI by our proposed method is always higher than ARI of other clustering methods for any $k_2$.
From Figs.~\ref{fg: lambda} and \ref{fg: k2}, we confirmed that ARI is stable against the selection of $\lambda$ (when $\lambda<0.1$) and $k_2$ in a wide range.
Thus, in practice, we suppose \proposal would perform well when $\lambda$ and $k_2$ may be simply chosen e.g., $\lambda=0.01$ and $k_2=k_1$.

As for the hyperparameter of PU learning, $\rho$ has a large influence on the performance of \proposal as shown in Figs.~\ref{fg: lambda} and \ref{fg: k2}.
To evaluate the effectiveness of the Positive Unlabeled approach, we show the effect of $\rho$ to ARI achieved by \proposal in Table~\ref{tb: q-ari}.
It shows that, when the density is high, the best $\rho$ tends to be low in general.

\subsection{Efficiency}
Table~\ref{tb: runtime} shows the runtimes of \proposal, JWNMF, and BAGC.
The hyperparameter $k_2$ is set to $k_1$ in our method.
First, we investigate the performance of our method without PU learning by setting $\rho$ to $0.5$ (Prop. (w/o PU) in the table).
In this case, the first term (topology part) of the loss function Eq.(\ref{eq: objective of proposal}) corresponds to $\frac{1}{2}\|\bmS-\bmU\bmU\trans\|^2_\F$, 
which is equivalent with the topology part of the JWNMF objective function Eq.(\ref{eq: objective of jwnmf}).
So, the runtime difference between Prop. (w/o PU) and JWNMF is caused by the different cost for the attribute part. 
The results in the table show that Prop. (w/o PU) is more efficient than JWNMF in all the datasets.
This is because our method is more efficient for the learning cost for the attribute part.

Next, we investigate the cost of PU learning.
Our method with PU learning (Prop.) requires more running time than without PU learning, Prop. (w/o PU). 
This is because the update rule for PU learning needs multiple times of the computation ($O(n^2kt)$) of the topology part. 

Finally, we compared the performance of our method with others.
Remember the discussion we made on the computational complexity.
Our method is more efficient than JWNMF by $O((m^2)kt)$.
So, it is expected that our method is faster than JWNMF for the datasets if its number of attributes ($m$) is large. 
Indeed, the results show that our method is more efficient than JWNMF on WebKB and Citeseer, which have many attributes and a  relatively small number of vertices.
As for BAGC, it is highly efficient but the clustering quality is very poor.
Thus, \proposal achieves high efficiency and effectiveness at the same time.



\input{table/runtime}

%% file: fig/0_visual.tex
\begin{figure}[t]
\centering
\begin{minipage}{0.36\hsize}
 \centering
  \includegraphics[trim={10mm 0 15mm 0},width=3.0cm]{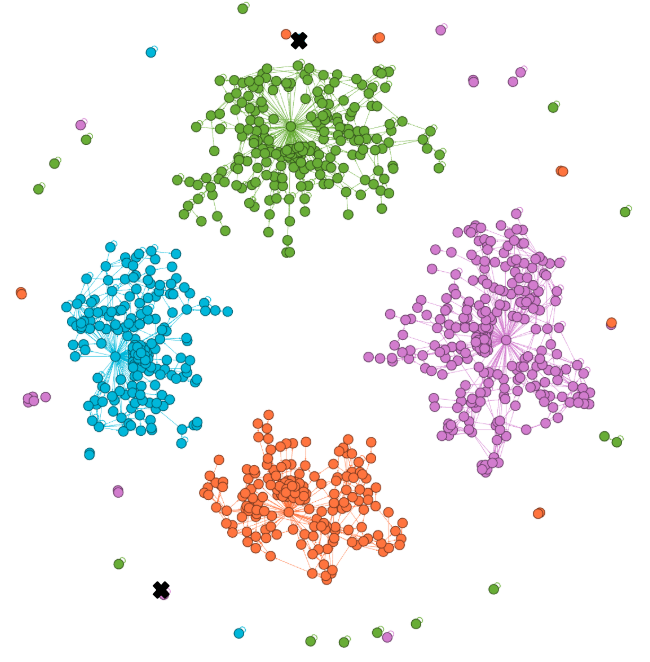}
  \subcaption{Proposed
  }
  \label{fg: WebKB_njnmf}
\end{minipage}
\hspace{10mm}
\centering
\begin{minipage}{0.36\hsize}
  \centering
  \includegraphics[trim={10mm 0 15mm 0},width=3.0cm]{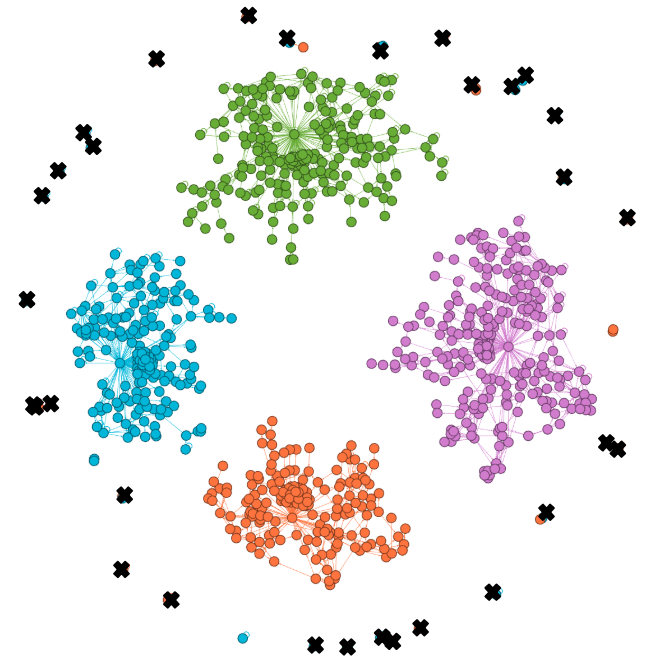}
  \subcaption{JWNMF}
  \label{fg: WebKB_jwnmf}
\end{minipage}
\caption{Visualization of the results of WebKB with four clusters. The colors of the vertices correspond to the clusters computed by our proposed method and JWNMF. The vertices with the cross mark (``x") indicate the wrong cluster assignment based on the ground truth. Our method assigns only two vertices to the wrong clusters out of $877$ vertices.}
  \label{fg: visualization}
\end{figure}

%% file: table/dataset.tex
\begin{small}
\begin{table}[t]
\setlength{\tabcolsep}{5pt}
  \caption{Summary of the datasets. }
  \label{tb: dataset}
  \begin{center}
	\begin{tabular}{|l|r|r|r|r|r|} \hline
		Dataset & Vertex & Edge & Attribute & Label & Density \\ 
		        & \multicolumn{1}{c|}{$n$} & \multicolumn{1}{c|}{$|E|$} & \multicolumn{1}{c|}{$m$} & \multicolumn{1}{c|}{$k_1$} & \multicolumn{1}{c|}{$|E|/n^2$} \\ \hline \hline
        WebKB & $877$ & $1480$ & $1703$ & $4$ &$0.18\%$\\ \hline
		Citeseer & $3312$ & $4660$ & $3703$ & $6$ & $0.04\%$\\ \hline
        Cora & $2708$ & $5278$ & $1433$ & $7$ &$0.07\%$ \\ \hline
        polblog & $1490$ & $16630$ & $7$ & $2$ & $0.75\%$\\ \hline
	\end{tabular}
 \end{center}
\end{table}
\end{small}

%% file: table/ari.tex
\begin{small}
\begin{table*}[t]\setlength{\tabcolsep}{5pt}
  \caption{The average and standard deviation (in parenthesis) of ARI. The methods annotated with * indicate the parameters are initialized by random values. 
  The boldface font represents the best performance for each dataset.}
  \label{tb: ari}
  \setlength{\tabcolsep}{5pt}
  \begin{center}
	\begin{tabular}{|l|l|c|c|c|c|} \hline
        Method & Input Type & WebKB & Citeseer & Cora  &  polblog \\ \hline \hline
        Prop. & Topology, Attribute & ${\bf 0.995} ~{(\pm 0.002)}$ &$ {\bf 0.280} ~{(\pm 0.027)} $& ${\bf 0.348} ~{(\pm 0.022)}$ &${\bf 0.626} ~{(\pm 0.037)}$\\ \hline
        Prop. (w/o PU)&Topology, Attribute &$0.990~{(\pm 0.005)}$&$0.221~{(\pm 0.010)}$&$0.270~{(\pm 0.024)}$&$0.621~{(\pm 0.000)}$\\ \hline
        Prop. $^{*}$& Topology, Attribute &$ 0.982~{(\pm 0.003)}$ & $ 0.126 ~{(\pm 0.023)}$ & $0.244 ~{(\pm 0.038)}$ &  $ 0.603 ~{(\pm 0.011)}$\\ \hline
        JWNMF  & Topology, Attribute &$0.906 ~{(\pm 0.000)}$ & $0.127 ~{(\pm 0.000)}$ & $0.230 ~{(\pm 0.000)}$ & $0.517 ~{(\pm 0.000)}$ \\ \hline
        JWNMF$^{*}$ & Topology, Attribute &$0.909 ~{(\pm 0.002)}$ &$ 0.082 ~{(\pm 0.009)}$ & $ 0.227 ~{(\pm 0.011)}$ & $0.504 ~{(\pm 0.011)}$\\ \hline
        BAGC & Topology, Attribute &$0.204 ~{(\pm 0.000)}$ & $0.000 ~{(\pm 0.000)}$ & $0.016 ~{(\pm 0.000)}$ & $0.000 ~{(\pm 0.000)}$\\ \hline
        METIS & Topology &$0.851 ~{(\pm 0.000)}$ & $0.156 ~{(\pm 0.000)}$ & $0.283 ~{(\pm 0.000)}$ & $0.545 ~{(\pm 0.000)}$\\ \hline 
        SNMF & Topology &$0.840 ~{(\pm 0.100)}$ &$ 0.067 ~{(\pm 0.020)}$ & $0.211 ~{(\pm 0.023)}$ &  $0.498 ~{(\pm 0.059)}$ \\ \hline
        NMF & Attribute &$0.327 ~{(\pm 0.004)}$ & $0.193 ~{(\pm 0.023)}$ & $0.115 ~{(\pm 0.001)}$ & $0.000 ~{(\pm 0.000)}$  \\ \hline
        k-means & Attribute &$0.260 ~{(\pm 0.131)}$ &$  0.190 ~{(\pm 0.044)}$ & $0.093 ~{(\pm 0.034)}$ & $0.000 ~{(\pm 0.000)}$\\ \hline 
 	\end{tabular}
 \end{center}
\end{table*}
\end{small}

%% file: table/modularity.tex
\begin{table}[t]
  \caption{Modularity and average entropy results for WebKB dataset. 
  The methods with * indicate the parameters are initialized by random values.
  The boldface font represents the best performance for each measure. 
  }
  \label{tb: modularity}
  \setlength{\tabcolsep}{5pt}
  \begin{center}
	\begin{tabular}{|l|r|r|c|} \hline
		Measure & Modularity & Entropy& ARI \\  \hline \hline
        Prop. & $0.738$ & $0.152$ &${\bf 0.995}$ \\ \hline
        Prop. (w/o PU)&$0.737$&$0.152$&$0.990$ \\ \hline
        Prop.$^{*}$ & $0.737$& $0.152$ &$0.982$ \\\hline
        JWNMF & $0.739$ & $0.153$ &$0.906$  \\ \hline
		JWNMF$^{*}$ & ${\bf 0.741}$& $0.153$ & $0.909$ \\ \hline
        BAGC & $0.224$ & $0.150$ & $0.204$\\ \hline
        METIS & $0.732$ & $0.153$ & $0.851$ \\ \hline 
        SNNF & $0.725$ & $0.153$ & $0.840$\\ \hline
        NMF & $0.278$ & ${\bf 0.146}$ & $0.327$ \\ \hline
        k-means & $0.239$ & ${\bf 0.146}$ & $0.260$ \\ \hline 
 	\end{tabular}
 \end{center}
\end{table}

%% file: fig/0_lambda.tex
\begin{figure}[t]
\centering
 \begin{minipage}{0.45\hsize}
  \centering
  \includegraphics[trim={10mm 0mm 0mm 0mm},width=4.4cm]{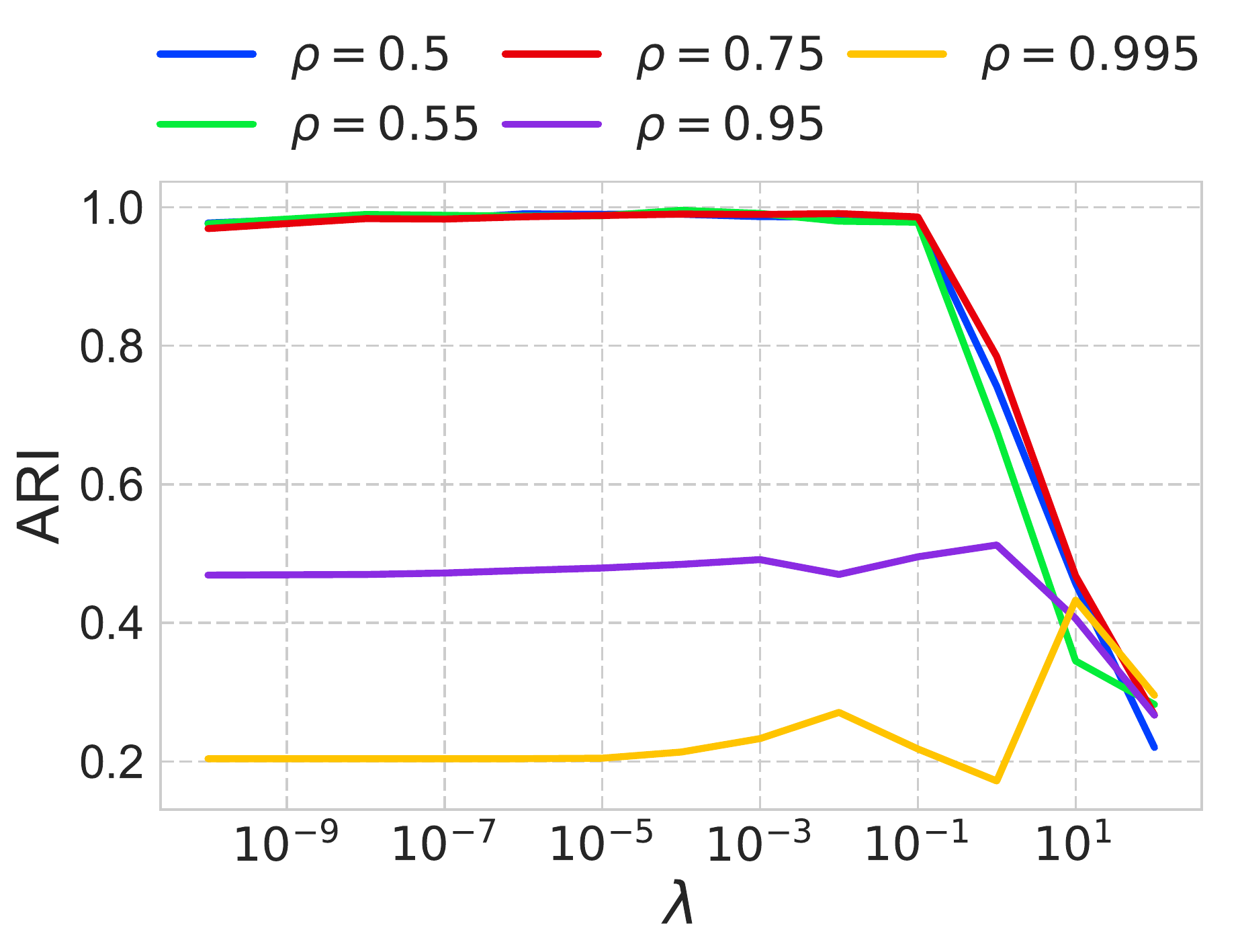}
  \subcaption{WebKB}
  \label{fg: WebKB_lambda}
  \end{minipage}
 \hspace{0.6cm}
 \begin{minipage}{0.45\hsize}
  \centering
  \includegraphics[trim={10mm 0mm 0mm 0mm},width=4.4cm]{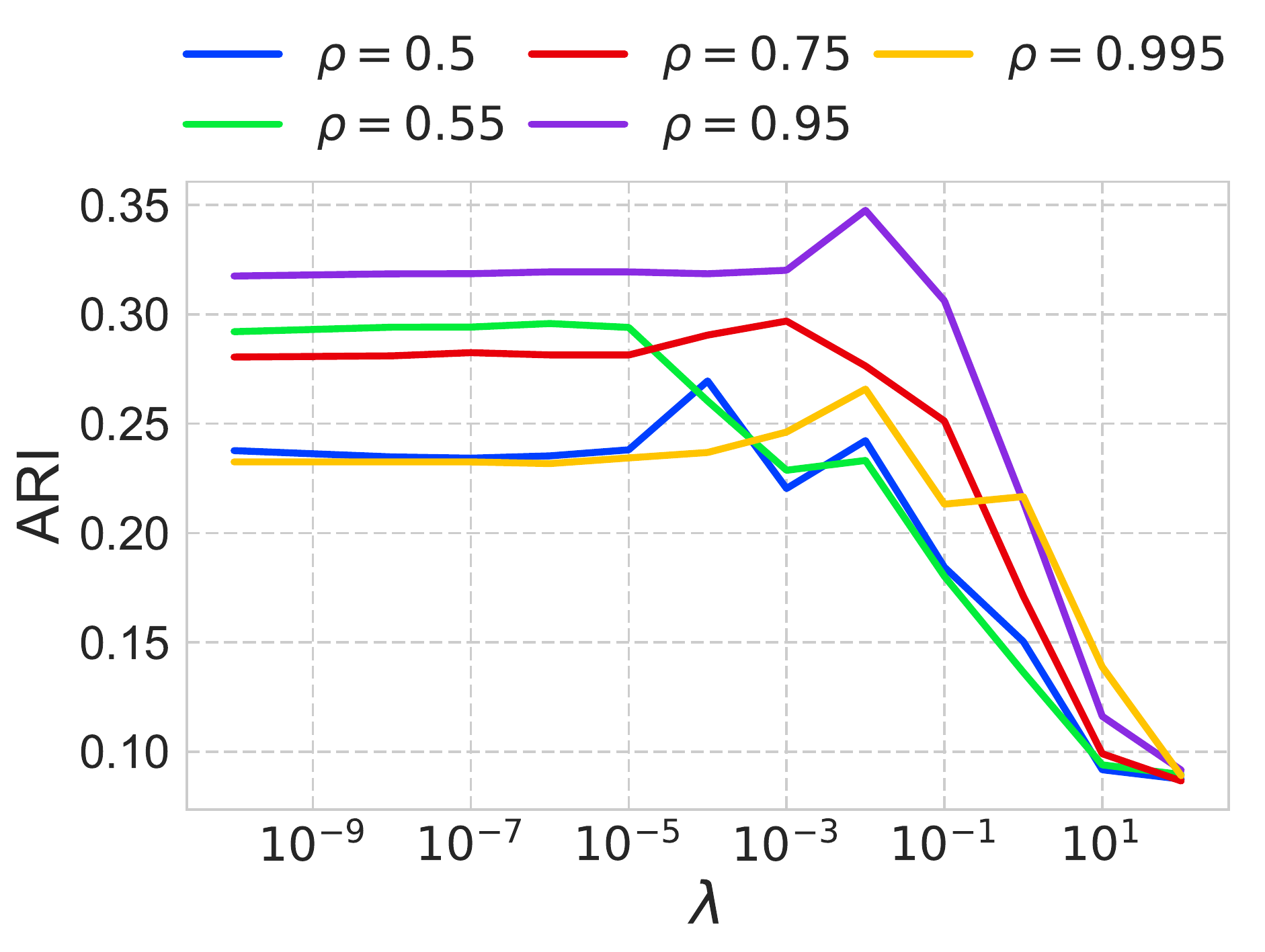}
  \subcaption{Cora}
  \label{fg: cora_lambda}
  \end{minipage}
\caption{Effect of $\lambda$ on clustering quality ARI in \proposal with k-means initialization for two datasets.}
  \label{fg: lambda}
\end{figure}

%% file: fig/0_k2.tex
\begin{figure}[t]
\centering
 \begin{minipage}{0.45\hsize}
  \centering
  \includegraphics[trim={10mm 0mm 0mm 0mm},width=4.4cm]{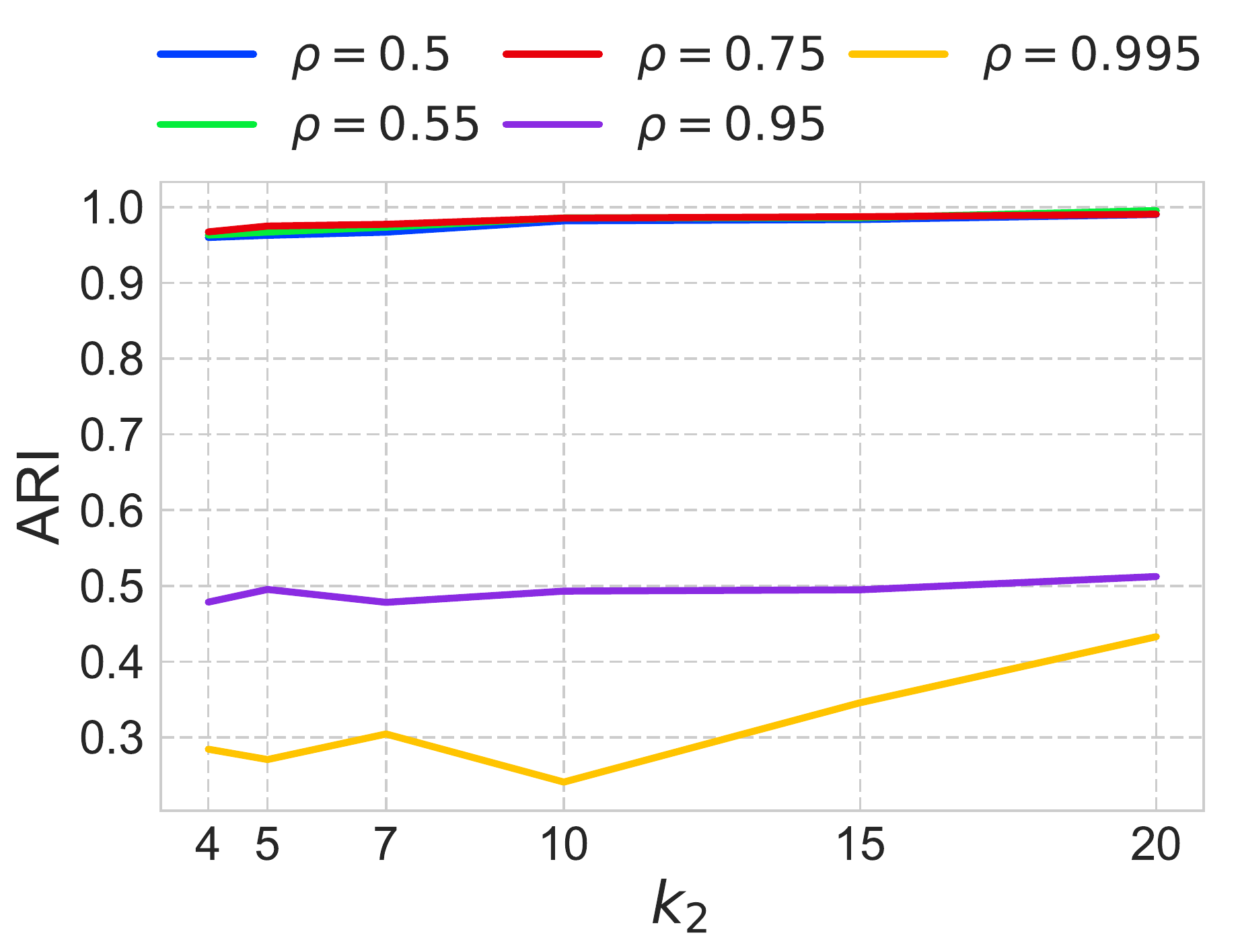}
  \subcaption{WebKB}
  \label{fg: WebKB_k2}
  \end{minipage}
  \hspace{0.6cm}
 \begin{minipage}{0.45\hsize}
  \centering
  \includegraphics[trim={10mm 0mm 0mm 0mm},width=4.4cm]{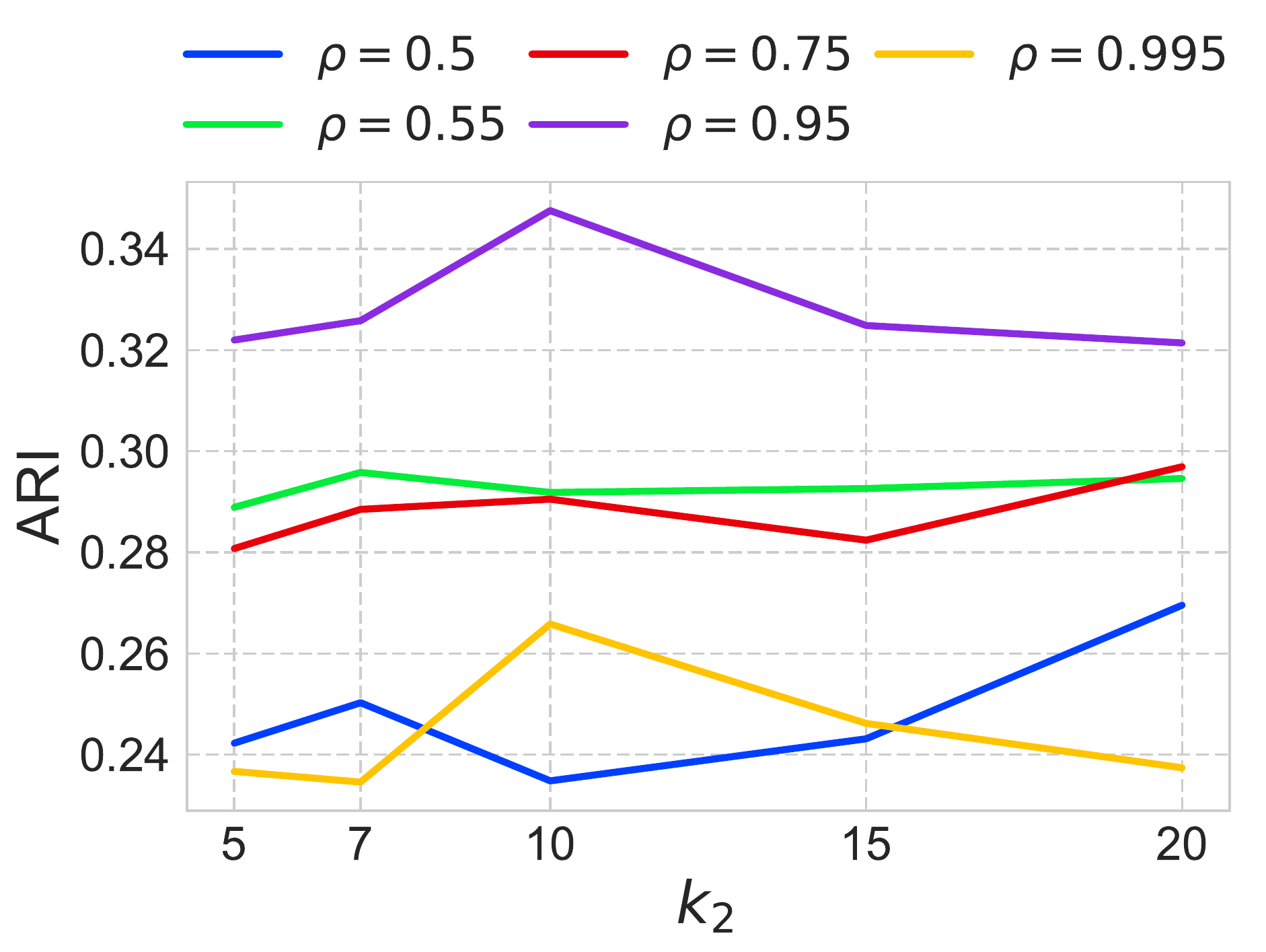}
  \subcaption{Cora}
  \label{fg: cora_k2}
  \end{minipage}
\caption{Effect of $k_2$ on clustering quality ARI in \proposal with k-means initialization for two datasets.}
  \label{fg: k2}
\end{figure}

%% file: table/q-ari.tex
\begin{table}[t]
  \caption{Effect of $\rho$ on ARI achieved by \proposal with k-means initialization. 
  The density indicates the (\# of observed edges)/(\# of possible edges), that is ${|{E}|/\ n^2}$.
  The boldface font represents the best performance for each dataset.
  }
  \label{tb: q-ari}
  \setlength{\tabcolsep}{5pt}
  \begin{center}
	\begin{tabular}{|l|r|r|r|r|} \hline
Dataset & WebKB & Citeseer & Cora & polblog \\ \hline \hline
Density &  0.18 \% & 0.04 \% & 0.07\% & 0.75 \% \\\hline \hline
$\rho=0.5$ & $0.990$ & $0.221$ & $0.270 $& $0.621$ \\ \hline
$\rho=0.55$ & ${\bf 0.995}$ & $0.216$ & $0.296$ & $0.625$ \\ \hline
$\rho=0.75$ & $0.991$ & $0.229$ & $0.297$ & $0.625$ \\ \hline
$\rho=0.95$ & $0.512$ & $0.254$ & ${\bf 0.348}$ & ${\bf 0.626}$ \\ \hline
$\rho=0.995$ & $0.433$ & ${\bf 0.280}$ & $0.266$ & $0.529$ \\ \hline
  \end{tabular}
  \end{center}
\end{table}

%% file: table/runtime.tex
\begin{table}[t]
  \caption{Runtime [sec] of attributed graph clustering methods. Our method and JWNMF are implemented in Python$3$ and BAGC is implemented in Octave.}
  \label{tb: runtime}
\setlength{\tabcolsep}{5pt}
  \begin{center}
	\begin{tabular}{|l|r|r|r|r|} \hline
	Method & WebKB & Citeseer & Cora & polblog\\ \hline \hline
        Prop. & 2.62 & 54.32 & 29.13 & 5.43\\ \hline
        Prop. (w/o PU) & 0.81 & 10.15 & 4.82 & 0.40\\ \hline
		JWNMF & 8.80& 60.71 & 13.33 & 0.60\\ \hline
		BAGC & 4.20 & 9.38 & 4.84 & 0.06\\ \hline
 	\end{tabular}
 \end{center}
\end{table}

%% file: 06_related.tex
\section{Related Work}
\label{sec:related}
SNMF is recently extended to consider both a graph structure and attribute information for discovering clusters of data entities.
JWNMF~\cite{huang2017joint} factorizes both the topology and the attribute matrices at the same time. The objective is shown:
\begin{align}
\label{eq: objective of jwnmf}
\min_{\bmU, \bmV, \bmA \ge 0} \|\bmS-\bmU\bmU\trans\|^2_F + \lambda\|\bmX \bmA - \bmU\bmV\trans\|^2_F
\end{align}
where $\lambda$ and $\bmA = {\rm diag}(a_1,a_2,\dots,a_k)$ are a hyper parameter and an attribute selection matrix, respectively.
There are two critical issues in JWNMF.
The first issue is that the clustering quality is not high. 
JWNMF is limited to learn the single cluster assignment matrix $\bmU$. Thus it does not fully capture the relationship between the topology and the attributes. 
The second issue is on the efficiency. JWNMF uses two model parameters $\lambda$ and ${\bmA}$ for adjusting attributes weight to the learning model, but they are redundant and, moreover, the cost of learning ${\bmA}$ is expensive as O($m^2$) where $m$ is the number of attributes. 

SA-Cluster~\cite{zhou2009graph} and its efficient version Inc-Cluster~\cite{zhou2010clustering} are the attributed graph clustering methods expanded from distance-based graph clustering.
The key idea is to embed vertex attributes as new vertices into the graph. 
A unified distance for the augmented graph is defined by the random walk process, and the graph is partitioned by k-medoids.
It is hard to apply these methods to large graphs since the augmented steps increase the size of the graph considerably.

BAGC/GBAGC~\cite{xu2012model,Xu2014GBAGCAG} learns a posterior distribution over the model parameters.
This method assumes that the vertices in the same cluster should have a common multinomial distribution for each vertex attribute and a Bernoulli distribution for vertex connections. 
The attributed graph clustering problem can be solved as a standard probabilistic inference problem.

PAICAN~\cite{bojchevski2018bayesian} performs anomaly detection and clustering on the attributed graph at the same time.
PAICAN explicitly models partial anomalies by generalizing ideas of Degree Corrected Stochastic Block Models~\cite{karrer2011stochastic,yan2014model} and Bernoulli Mixture Models.
This method achieves high clustering quality after removing anomalies it detects. 
The drawback of PAICAN is that it can only handle categorical attributes.

Graph Convolutional Networks~~\cite{kipf2017semi}, that is a semi-supervised learning method for a graph, has obtained considerable attention from machine learning and data mining fields due to its high performance in classifying graph vertices. However, this approach needs a subset of true cluster labels on vertices, and thus its goal is different from that of the attributed graph clustering.

%% file: 05_conclusion.tex
\section{Conclusion}
\label{sec:conclusion}
We considered the clustering problem of attributed graphs.
We designed an effective and efficient clustering method, \nagc, Non-linear Attribute Graph Clustering by Symmetric Non-negative Matrix Factorization with Positive Unlabeled Learning.
The features of \proposal are three holds. 
1) it learns a non-linear projection between the two latent embedding spaces of the topology and the attributes of graphs,
2) it leverages the positive unlabeled learning to take the effect of partially observed positive edges, and
3) it achieves efficient computational complexity, $O((n^2+mn)kt)$ for learning the \cls.